\pdfoutput=1

\documentclass[11pt]{article}

\usepackage{acl}

\usepackage{times}
\usepackage{latexsym}

\usepackage[T1]{fontenc}

\usepackage[utf8]{inputenc}

\usepackage{microtype}

\usepackage{inconsolata}

\usepackage{booktabs} %
\usepackage{listings} %
\usepackage{tikz} %
\usepackage{pgfplots} %
\pgfplotsset{compat=1.17} %

\usepackage[most]{tcolorbox} %

\usepackage[normalem]{ulem}  %
\usepackage{xcolor}
\newcommand{\magentauline}[1]{{\color{magenta}\uline{{\color{black}#1}}}}

\title{Distilling Named Entity Recognition Models for Endangered Species \\ from Large Language Models}

\author{Jesse Atuhurra  \\ \And
  Seiveright Cargill Dujohn\\ \And
    Hidetaka Kamigaito\\ \AND
  Hiroyuki Shindo\\ \And
  Taro Watanabe\\ \AND
  \texttt{Division of Information Science, NAIST} \\ 
  \footnotesize
  {
  \texttt{ \{atuhurra.jesse.ag2, seiveright.cargill\_dujohn.sf4, kamigaito.h, shindo, taro\} @naist.ac.jp} 
  } \\ }

\begin{document}
\maketitle
\begin{abstract}
Natural language processing (NLP) practitioners are leveraging large language models (LLM) to create structured datasets from semi-structured and unstructured data sources such as patents, papers, and theses, without having domain-specific knowledge. At the same time, ecological experts are searching  for a variety of means to preserve biodiversity. To contribute to these efforts, we focused on endangered species and through in-context learning, we distilled knowledge from GPT-4~\cite{openai2023gpt4}. In effect, we created datasets for both named entity recognition (NER) and relation extraction (RE) via a two-stage process: 1) we generated synthetic data from GPT-4 of four classes of endangered species, 2) humans verified the factual accuracy of the synthetic data, resulting in gold data. Eventually, our novel dataset contains a total of 3.6K sentences, evenly divided between 1.8K NER and 1.8K RE sentences. The constructed dataset was then used to fine-tune both \textit{general BERT} and \textit{domain-specific BERT} variants, completing the knowledge distillation process from GPT-4 to BERT, because GPT-4 is resource intensive. Experiments show that our knowledge transfer approach is effective at creating a NER model suitable for detecting endangered species from texts. 
\end{abstract}

\section{Introduction}
\label{sec:Introduction}
\begin{figure}[!t]
\centering
\includegraphics[width=7.5cm, height=5cm]{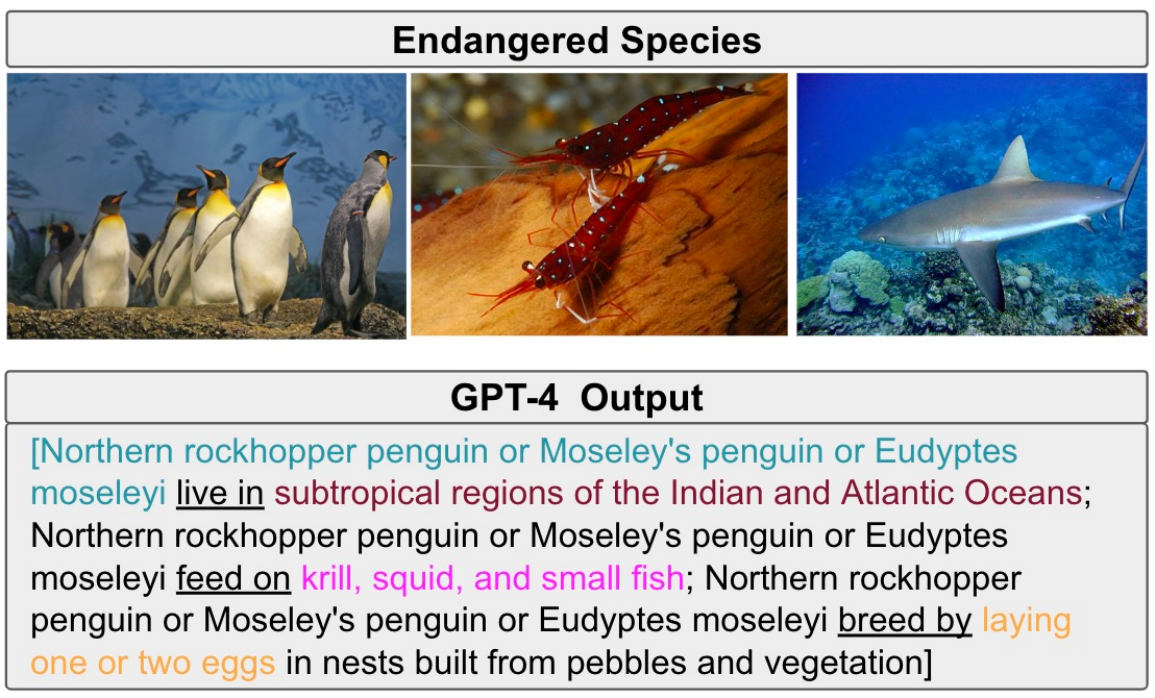}
\caption{Illustration of GPT-4 NE and relations for a unique species. We created NER data for four named entities; \texttt{species, habitat, feeding, breeding}, and RE data with three relation classes; \texttt{live\_in, feed\_on, breed\_by}}
\label{fig:IntroGPT4Prompt}
\end{figure}
Natural language processing (NLP) practitioners are leveraging large language models (LLM) to create structured datasets from semi-structured and unstructured data sources, such as patents, papers and theses, without having domain-specific knowledge. At the same time, ecological experts are searching  for a variety of means to preserve biodiversity because critical endangerment and extinction of species
can drastically alter biodiversity, threaten the global ecology, and negatively impact the livelihood of people~\cite{do2020research}.
Information about species are often stored in scientific literature in the form of free flowing natural language that is not readily machine parsable \cite{swain2016chemdataextractor}. These scientific works store latent information that are not leveraged for advanced machine learning discoveries \cite{dunn2022structured}. 
Hence, there is a surge of demand to convert scientific works into structured data by researchers~\cite{gutierrez2022thinking}. 
To contribute to these efforts, in this study, we focused on endangered species to capture the interactions between species, their trophic level, and habitat~\cite{christin2019applications}. We distilled knowledge from GPT-4~\cite{openai2023gpt4} via in-context learning~\cite{NEURIPS2020_1457c0d6}. We created NER and RE datasets via a two-stage process: 1) we generated synthetic data from GPT-4 of four classes of endangered species, namely, amphibians, arthropods, birds, fishes, 2) humans verified the factuality of the synthetic data, resulting in gold data. Eventually, our novel dataset contains 3.6K sentences, evenly divided between 1.8K NER and 1.8K RE sentences. 
\begin{figure*}[!t]
\centering
\includegraphics[width=0.68\linewidth]{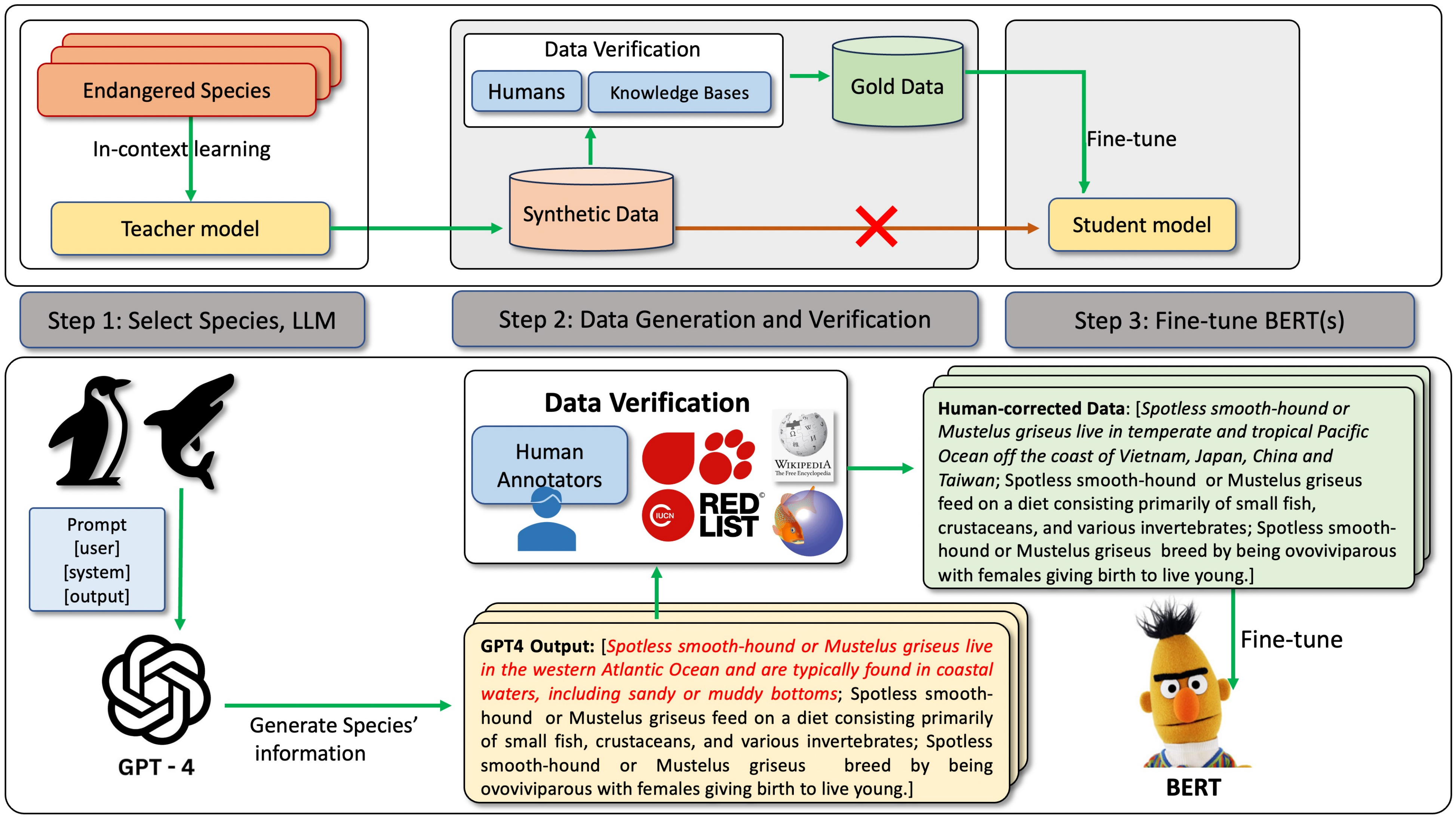} %
\caption{
Steps involved in the transfer of knowledge from GPT-4 (teacher) to BERT (student). When, GPT-4 output is incorrect (text shown in red), humans corrected the data. We leveraged external knowledge from knowledge bases such as IUCN, Wikipedia, FishBase, and more, to verify all the species' data. Lastly, we fine-tuned BERT variants. 
}
\label{fig:Project_Overview}
\end{figure*}
The new dataset was then used to fine-tune both \textit{general BERT} and \textit{domain-specific BERT} variants, completing the knowledge distillation process from GPT-4 to BERT, because GPT-4 is resource intensive. Experiments show that our knowledge transfer approach is effective at creating a NER model suitable for detecting endangered species from texts. 
Moreover, further human evaluation for zero-shot NER with both GPT-4 and UniversalNER\footnote{UniversalNER-7B is a LLM developed specifically for NER, and is available here \url{https://huggingface.co/Universal-NER/UniNER-7B-all}}\cite{zhou2023universalner} reveal that GPT-4 is a good teacher model. 
\section{Knowledge Distillation}
\label{sec:KnowledgeDistillation}
Despite the impressive performance of LLM, they are resource intensive and closed-source, harboring concerns about privacy and transparency. Moreover, they are costly to use whether through running these models in-house or accessing their APIs via subscription \cite{brown2020language, zhou2023universalner, agrawal2022large, wang2021want}. 

Knowledge distillation has shown to circumvent these challenges while maintaining or even surpassing the performance of large models. \newcite{hinton2015distilling, wang2021want, liu2019improving} proposed strategies to distill complex models into smaller models for downstream tasks. Furthermore, studies by \newcite{wang2021want, lang2022cotraining, smith2022language} demonstrated that \texttt{prompting+resolver} can outperform LLM. In particular, the pipeline from \cite{ratner2017snorkel} was leveraged to collect LLM-generated outputs to train a smaller task-specific model on CASI through weak supervision \cite{agrawal2022large}.

In short, knowledge distillation allows for the transfer of knowledge from large models to smaller models for many downstream tasks \cite{hinton2015distilling, wang2021want}, overcoming challenges associated with LLM.

\section{Dataset Creation}
\label{sec:DatasetCreation}
Dataset creation is shown in Figure~\ref{fig:Project_Overview}. First, we applied prompts in GPT-4 to generate  data for all species (in step 1\&2). Then, all of this \textit{synthetic data} was verified by humans (in step 2). The verified data is the \textit{gold data}.

\subsection{Endangered Species}
\label{subsec:EndangeredSpecies}
In order to test our hypothesis, we chose the bio domain and focused on endangered species\footnote{The list of Endangered Species is available at \url{https://en.wikipedia.org/wiki/Lists_of_IUCN_Red_List_endangered_species}. This list is officially maintained by The International Union for Conservation of Nature (IUCN) who regularly update information regarding threats to species' existence. The list is dabbed \textit{Red List} and can be found here \url{https://en.wikipedia.org/wiki/IUCN_Red_List}.} %
All the species studied in this work have a Wikipedia page dedicated to them. This requirement allowed us to minimize difficulty in finding  information relevant to verify the data generated by GPT-4. 

We investigated four classes of species: \texttt{amphibians, arthropods, birds, fishes}. For each class, we collected data of 150 unique species. Moreover, due to the scientific importance of \texttt{common names} and \texttt{scientific names} for each species, we mandated that all sentences contained in our dataset carry both names. Sentence format: \texttt{[common name] or [scientific name] live in; } (illustrated in Table~\ref{Table: Annotation process for NEs}).

\subsection{In-context Learning with GPT-4}
\label{subsec:ICLwithGPT4}
After deciding the categories, we distilled knowledge from GPT-4\footnote{Our study is based on the GPT-4 version available in May 2023 on the ChatGPT user interface.} about each unique species. 
We leverage in-context learning and apply prompts to GPT-4 to generate data regarding the species' \textit{habitat, feeding, breeding}.
In short, GPT-4 generated three sentences describing the habitat, feeding, and breeding for each species, contained in one tuple. We refer to the generated data as \textit{synthetic data}. The prompt is shown in Table \ref{Table:InputPrompt}.
\begin{table}[!t]
\footnotesize
\centering
\begin{tabular}{l}
\toprule
\textbf{Input Prompt} \\
\midrule
A habitat provides the necessary resources for survival,\\
such as food, water, shelter, and space for breeding and \\
movement for any particular plant, animal, or organism. \\
Let us define a new variable, i.e, \\
species ==  Northern rockhopper penguin. Where does \\
the species live, what does species feed on, and how does \\
species breed? Give answer as a tuple in this format: \\
(species lives in..; species feeds on..; species breeds by..) \\
\bottomrule
\end{tabular}
\caption{\label{Table:InputPrompt} Prompt used to generate data. The full prompt is shown in Appendix~\ref{Appendix:GPT4InputPrompt}.}
\end{table}

Due to the hallucination-nature of LLM, GPT-4 often generated incorrect species information. Human annotators helped with the verification of all GPT-4 data. 

\subsection{Data Verification}
The need to correct the synthetic data led to a robust data verification process. %
The time needed to verify the factual accuracy of GPT-4 text for NE and relations of one species varied between 5 minutes and several hours. The data verification process results into the \textit{gold data}. 

There are two major components of this process; 1) knowledge bases (KB) which provide the reliable external knowledge relevant to establish the correctness of new sentences from GPT-4. KB used in this study include: IUCN\footnote{The official IUCN page can be found here~\url{https://www.iucnredlist.org/}}, Wikipedia, FishBase\footnote{This knowledge base provides information about fish species. URL~\url{https://www.fishbase.se/search.php}}, and more. Then, 2) humans read each new sentence and with the help of the above KB, human annotators confirmed if the information provided by GPT-4 about each species' habitat, feeding, and breeding were correct or not. Whenever such information was false, humans manually corrected the sentences. Table~\ref{Table:GPT4FactualAccuracy} summarizes the quality of GPT-4 data for each named entity (NE). More details in Appendix~\ref{Appendix:GPT4Quality}.
\begin{table}[!t]
\footnotesize
\centering
\begin{tabular}{llll}
\toprule
\textbf{Entity} & Breeding   & Feeding   & Habitat \\
\midrule
\textbf{F1 (\%)} &  74.14    & 75.35     & 73.26 \\
\bottomrule
\end{tabular}
\caption{\label{Table:GPT4FactualAccuracy} Factual correctness of data generated by GPT-4, measured by F1. The \textbf{average-F1} is 74.25\%. }
\end{table}

\subsection{NER and RE Data}
\label{subsec:NERandREData}
In order to obtain the data necessary to fine-tune BERT and its domain-specific variants for NER and RE, the verified sentences were annotated as follows. For NER, we adopt the \textit{CoNLL format} in which one column contains \textit{tokens} and the other column contains the \textit{BIO tags}. These are the four named entities in our data; \texttt{SPECIES, HABITAT, FEEDING, BREEDING}. An annotated NER example is shown in Table~\ref{Table: Annotation process for NEs}. For the RE data, we defined three classes of relations, namely; \texttt{live\_in, feed\_on, breed\_by}, to describe the specie's habitats, feeding behavior, and reproduction process, respectively. We followed the format introduced by \newcite{baldini-soares-etal-2019-matching}.
\begin{table}[!t]
\footnotesize
\centering
\begin{tabular}{l}
\toprule
\textbf{Example of annotated NER sentences } \\
\hline
\magentauline{Smoothtooth blacktip shark}$^{\texttt{SPECIE}}$ or \\
\magentauline{Carcharhinus leiodon}$^{\texttt{SPECIE}}$ live \\
in \magentauline{warm coastal waters}$^{\texttt{HABITAT}}$ \\
particularly in the Indo-Pacific region; \\
\magentauline{Smoothtooth blacktip shark}$^{\texttt{SPECIE}}$ \\
or \magentauline{Carcharhinus leiodon}$^{\texttt{SPECIE}}$ \\
feed on \magentauline{small bony fish}$^{\texttt{FEEDING}}$, 
\magentauline{crustaceans}$^{\texttt{FEEDING}}$ \\
and \magentauline{cephalopods}$^{\texttt{FEEDING}}$; \\
\magentauline{Smoothtooth blacktip shark}$^{\texttt{SPECIE}}$ or \\
\magentauline{Carcharhinus leiodon}$^{\texttt{SPECIE}}$ breed by \\
\magentauline{giving birth to live shark pups}$^{\texttt{BREEDING}}$; \\
\bottomrule
\end{tabular}
\caption{\label{Table: Annotation process for NEs}
We annotated the entity mentions of \texttt{SPECIES, HABITAT, FEEDING, BREEDING} in each sentence.}
\end{table}

\subsection{Dataset Statistics}
\label{subsec:DatasetStatistics}
There are 1.8K new NER sentences. The NER data contains 607 unique species. In addition, there are 1.8K new RE sentences. The RE data contains; 607 \texttt{live\_in}, 582 \texttt{feed\_on}, and 570 \texttt{breed\_by} relations, respectively.
\section{Experiments}
\label{sec:Experiments}
The main goal of this study is to determine how effective is knowledge-transfer from teacher to student models, in extracting information about species from biological texts. We chose BERT and its variants, as students.
\subsection{General vs Domain-specific BERT}
\label{subsec:BERTfinetuning}
\paragraph{Models} We focused on the NER task, and chose three pre-trained models. Standard BERT-large\footnote{We adopt the pretrained bert-large-uncased available at \url{https://huggingface.co/bert-large-uncased}. }\cite{devlin-etal-2019-bert} is our general student model. We compared it with two models specific to the bio domain, namely, BioBERT-large\footnote{ We chose the biobert-large-cased-v1.1 version which is available here \url{https://huggingface.co/dmis-lab/biobert-large-cased-v1.1}. }\cite{Lee_2019}, and PubMedBERT-large\footnote{Please note that PubMedBERT-large has a new name, BiomedNLP-BiomedBERT-large-uncased-abstract, and it is available at \url{https://huggingface.co/microsoft/BiomedNLP-BiomedBERT-large-uncased-abstract}. }\cite{pubmedbert}.

The three models were fully fine-tuned on the novel data, to complete the knowledge distillation process from GPT-4. During fine-tuning, we ran each experiment two times with different seeds for 20 epochs, and reported the average scores.

\paragraph{Results}
Table \ref{Table:F1scoresNamedEntity} shows the average F1-score per NE for all student models.
BERT, BioBERT, and PubMedBERT achieve competitive F1-scores, indicating that students learned to detect entity information relevant to endangered species. Indeed, our student models surpassed the teacher model, GPT-4. PubMedBERT outperforms GPT-4 by \textbf{+19.89\%} F1-score.

\begin{table}[!t]
\footnotesize
\centering
\begin{tabular}{llrrr}
\toprule
\textbf{Entity} & \textbf{BERT} & \textbf{BioBERT}  & \textbf{PubMedBERT} \\
\midrule
Breeding    & 94.65 & 94.26 & 95.78 \\
Feeding     & 91.49 & 93.29 & 90.26 \\
Habitat     & 87.54 & 87.36 & 90.97 \\
Species      & 99.39 & 99.25 & 99.46 \\
\midrule
\textbf{Average-F1} & \textbf{93.27} & \textbf{93.54} & \textbf{94.14} \\
\bottomrule
\end{tabular}
\caption{\label{Table:F1scoresNamedEntity} F1-score (\%) for each NE and average performance of all student models across all NE. PubMedBERT performs better than both BERT and BioBERT.}
\end{table}
\section{Discussion}
\subsection{Is Data Verification Effective?}
\label{subsec:EffectiveVerification }
After evaluating the quality of data generated by GPT-4, the average F1 is 74.25\%. By fine-tuning BERT and its variants on the human-verified data, F1 scores for all models are above 90\%. The results validate our efforts to verify the data, and also indicate that the student models learned to recognize NE about endangered species.
\subsection{Is GPT-4 a good teacher?}
\label{subsec:GPT4teacher}
To establish GPT-4's suitability as a teacher, we conducted a comprehensive analysis with zero-shot NER. We compared GPT-4 to a state-of-the-art NER-specific model, that is, \emph{UniversalNER-7B}. Both models were analysed by humans.
\paragraph{Human evaluation} We analysed the abilities of both LLM via human evaluation, and the analysis is two-fold. \textbf{First,} 100 samples were selected at random from the NER dataset and fed as input to both LLM. We measured how accurately the LLM extracted information from the text related to habitat, feeding and breeding for each species. We regard this evaluation as ``easy''. \textbf{Second,} we fed as input to both LLM more difficult text and again evaluated their zero-shot abilities. Here, difficult means that 3 to 5 paragraphs were fed to UniversalNER while longer text documents were fed to GPT-4 due to its much larger context window.  We refer to this evaluation as ``hard''. In both ``easy'' and ``hard'' evaluation settings above, we set the context length (that is, \textit{max\_length}) of \emph{Universal-NER/UniNER-7B-all} to 4,000 tokens.

As shown in Table~\ref{Table:HumanEvaluation}, GPT-4 is superior to \emph{UniversalNER-7B} at zero-shot NER, making it a suitable teacher model.
\begin{table}[!t]
\footnotesize
\centering
\begin{tabular}{lrr}
\toprule
\textbf{Text Input} & \textbf{GPT-4 }  & \textbf{UniversalNER-7B } \\
\midrule
Easy         & 100 & 78 \\
Hard          & 94 & 86 \\
\midrule
\textbf{Average-Acc} & \textbf{97} & \textbf{82} \\
\bottomrule
\end{tabular}
\caption{\label{Table:HumanEvaluation}Human evaluation of zero-shot NER for both GPT-4 and UniversalNER-7B on random samples of 100 ``easy'' and ``hard'' texts. We report the accuracy scores (see Appendix~\ref{Appendix:EasyAndHardExamples} for examples).} 
\end{table}
\section{Conclusion}
\label{sec:Conclusion}
In this study, we investigated the ability of LLM to generate reliable datasets suitable for training NLP systems for tasks such as NER. We constructed two datasets for NER and RE via a robust data verification process conducted by humans. The fine-tuned BERT models on our NER data achieved average F1-scores above 90\%. This indicates the effectiveness of our knowledge distillation process from GPT-4 to BERT, for NER in endangered species. We also confirmed that GPT-4 is a good teacher model.

\newpage
\bibstyle{acl_natbib} 
\bibliography{main}

\begin{thebibliography}{20}
\expandafter\ifx\csname natexlab\endcsname\relax\def\natexlab#1{#1}\fi

\bibitem[{Agrawal et~al.(2022)Agrawal, Hegselmann, Lang, Kim, and Sontag}]{agrawal2022large}
Monica Agrawal, Stefan Hegselmann, Hunter Lang, Yoon Kim, and David Sontag. 2022.
\newblock \href {http://arxiv.org/abs/2205.12689} {Large language models are few-shot clinical information extractors}.

\bibitem[{Baldini~Soares et~al.(2019)Baldini~Soares, FitzGerald, Ling, and Kwiatkowski}]{baldini-soares-etal-2019-matching}
Livio Baldini~Soares, Nicholas FitzGerald, Jeffrey Ling, and Tom Kwiatkowski. 2019.
\newblock \href {https://doi.org/10.18653/v1/P19-1279} {Matching the blanks: Distributional similarity for relation learning}.
\newblock In \emph{Proceedings of the 57th Annual Meeting of the Association for Computational Linguistics}, pages 2895--2905, Florence, Italy. Association for Computational Linguistics.

\bibitem[{Brown et~al.(2020{\natexlab{a}})Brown, Mann, Ryder, Subbiah, Kaplan, Dhariwal, Neelakantan, Shyam, Sastry, Askell, Agarwal, Herbert-Voss, Krueger, Henighan, Child, Ramesh, Ziegler, Wu, Winter, Hesse, Chen, Sigler, Litwin, Gray, Chess, Clark, Berner, McCandlish, Radford, Sutskever, and Amodei}]{NEURIPS2020_1457c0d6}
Tom Brown, Benjamin Mann, Nick Ryder, Melanie Subbiah, Jared~D Kaplan, Prafulla Dhariwal, Arvind Neelakantan, Pranav Shyam, Girish Sastry, Amanda Askell, Sandhini Agarwal, Ariel Herbert-Voss, Gretchen Krueger, Tom Henighan, Rewon Child, Aditya Ramesh, Daniel Ziegler, Jeffrey Wu, Clemens Winter, Chris Hesse, Mark Chen, Eric Sigler, Mateusz Litwin, Scott Gray, Benjamin Chess, Jack Clark, Christopher Berner, Sam McCandlish, Alec Radford, Ilya Sutskever, and Dario Amodei. 2020{\natexlab{a}}.
\newblock \href {https://proceedings.neurips.cc/paper_files/paper/2020/file/1457c0d6bfcb4967418bfb8ac142f64a-Paper.pdf} {Language models are few-shot learners}.
\newblock In \emph{Advances in Neural Information Processing Systems}, volume~33, pages 1877--1901. Curran Associates, Inc.

\bibitem[{Brown et~al.(2020{\natexlab{b}})Brown, Mann, Ryder, Subbiah, Kaplan, Dhariwal, Neelakantan, Shyam, Sastry, Askell et~al.}]{brown2020language}
Tom Brown, Benjamin Mann, Nick Ryder, Melanie Subbiah, Jared~D Kaplan, Prafulla Dhariwal, Arvind Neelakantan, Pranav Shyam, Girish Sastry, Amanda Askell, et~al. 2020{\natexlab{b}}.
\newblock Language models are few-shot learners.
\newblock \emph{Advances in neural information processing systems}, 33:1877--1901.

\bibitem[{Christin et~al.(2019)Christin, Hervet, and Lecomte}]{christin2019applications}
Sylvain Christin, Étienne Hervet, and Nicolas Lecomte. 2019.
\newblock \href {https://doi.org/10.1111/2041-210X.13256} {Applications for deep learning in ecology}.

\bibitem[{Devlin et~al.(2019)Devlin, Chang, Lee, and Toutanova}]{devlin-etal-2019-bert}
Jacob Devlin, Ming-Wei Chang, Kenton Lee, and Kristina Toutanova. 2019.
\newblock \href {https://doi.org/10.18653/v1/N19-1423} {{BERT}: Pre-training of deep bidirectional transformers for language understanding}.
\newblock In \emph{Proceedings of the 2019 Conference of the North {A}merican Chapter of the Association for Computational Linguistics: Human Language Technologies, Volume 1 (Long and Short Papers)}, pages 4171--4186, Minneapolis, Minnesota. Association for Computational Linguistics.

\bibitem[{Do et~al.(2020)Do, Choi, Hwang, Lee, Hur, Choi, Son, Kwon, Yoo, and Nam}]{do2020research}
Min~Su Do, Gabin Choi, Ji~Woo Hwang, Ji~Yeong Lee, Woo~Hyun Hur, Young~Su Choi, Seong~Ji Son, In~Kyeong Kwon, Seung~Youp Yoo, and Hyo~Kee Nam. 2020.
\newblock \href {https://doi.org/10.1016/j.japb.2020.09.008} {Research topics and trends of endangered species using text mining in korea}.

\bibitem[{Dunn et~al.(2022)Dunn, Dagdelen, Walker, Lee, Rosen, Ceder, Persson, and Jain}]{dunn2022structured}
Alexander Dunn, John Dagdelen, Nicholas Walker, Sanghoon Lee, Andrew~S. Rosen, Gerbrand Ceder, Kristin Persson, and Anubhav Jain. 2022.
\newblock \href {http://arxiv.org/abs/2212.05238} {Structured information extraction from complex scientific text with fine-tuned large language models}.

\bibitem[{Gu et~al.(2020)Gu, Tinn, Cheng, Lucas, Usuyama, Liu, Naumann, Gao, and Poon}]{pubmedbert}
Yu~Gu, Robert Tinn, Hao Cheng, Michael Lucas, Naoto Usuyama, Xiaodong Liu, Tristan Naumann, Jianfeng Gao, and Hoifung Poon. 2020.
\newblock \href {http://arxiv.org/abs/arXiv:2007.15779} {Domain-specific language model pretraining for biomedical natural language processing}.

\bibitem[{Gutierrez et~al.(2022)Gutierrez, McNeal, Washington, Chen, Li, Sun, and Su}]{gutierrez2022thinking}
Bernal~Jimenez Gutierrez, Nikolas McNeal, Clay Washington, You Chen, Lang Li, Huan Sun, and Yu~Su. 2022.
\newblock Thinking about gpt-3 in-context learning for biomedical ie? think again.
\newblock \emph{arXiv preprint arXiv:2203.08410}.

\bibitem[{Hinton et~al.(2015)Hinton, Vinyals, and Dean}]{hinton2015distilling}
Geoffrey Hinton, Oriol Vinyals, and Jeff Dean. 2015.
\newblock \href {http://arxiv.org/abs/1503.02531} {Distilling the knowledge in a neural network}.

\bibitem[{Lang et~al.(2022)Lang, Agrawal, Kim, and Sontag}]{lang2022cotraining}
Hunter Lang, Monica Agrawal, Yoon Kim, and David Sontag. 2022.
\newblock \href {http://arxiv.org/abs/2202.00828} {Co-training improves prompt-based learning for large language models}.

\bibitem[{Lee et~al.(2019)Lee, Yoon, Kim, Kim, Kim, So, and Kang}]{Lee_2019}
Jinhyuk Lee, Wonjin Yoon, Sungdong Kim, Donghyeon Kim, Sunkyu Kim, Chan~Ho So, and Jaewoo Kang. 2019.
\newblock \href {https://doi.org/10.1093/bioinformatics/btz682} {Biobert: a pre-trained biomedical language representation model for biomedical text mining}.
\newblock \emph{Bioinformatics}, 36(4):1234–1240.

\bibitem[{Liu et~al.(2019)Liu, He, Chen, and Gao}]{liu2019improving}
Xiaodong Liu, Pengcheng He, Weizhu Chen, and Jianfeng Gao. 2019.
\newblock \href {http://arxiv.org/abs/1904.09482} {Improving multi-task deep neural networks via knowledge distillation for natural language understanding}.

\bibitem[{OpenAI(2023)}]{openai2023gpt4}
OpenAI. 2023.
\newblock \href {http://arxiv.org/abs/2303.08774} {Gpt-4 technical report}.

\bibitem[{Ratner et~al.(2017)Ratner, Bach, Ehrenberg, Fries, Wu, and R{\'e}}]{ratner2017snorkel}
Alexander Ratner, Stephen~H Bach, Henry Ehrenberg, Jason Fries, Sen Wu, and Christopher R{\'e}. 2017.
\newblock Snorkel: Rapid training data creation with weak supervision.
\newblock In \emph{Proceedings of the VLDB Endowment. International Conference on Very Large Data Bases}, volume~11, page 269. NIH Public Access.

\bibitem[{Smith et~al.(2022)Smith, Fries, Hancock, and Bach}]{smith2022language}
Ryan Smith, Jason~A. Fries, Braden Hancock, and Stephen~H. Bach. 2022.
\newblock \href {http://arxiv.org/abs/2205.02318} {Language models in the loop: Incorporating prompting into weak supervision}.

\bibitem[{Swain and Cole(2016)}]{swain2016chemdataextractor}
Matthew~C Swain and Jacqueline~M Cole. 2016.
\newblock Chemdataextractor: a toolkit for automated extraction of chemical information from the scientific literature.
\newblock \emph{Journal of chemical information and modeling}, 56(10):1894--1904.

\bibitem[{Wang et~al.(2021)Wang, Liu, Xu, Zhu, and Zeng}]{wang2021want}
Shuohang Wang, Yang Liu, Yichong Xu, Chenguang Zhu, and Michael Zeng. 2021.
\newblock Want to reduce labeling cost? gpt-3 can help.
\newblock \emph{arXiv preprint arXiv:2108.13487}.

\bibitem[{Zhou et~al.(2023)Zhou, Zhang, Gu, Chen, and Poon}]{zhou2023universalner}
Wenxuan Zhou, Sheng Zhang, Yu~Gu, Muhao Chen, and Hoifung Poon. 2023.
\newblock \href {http://arxiv.org/abs/2308.03279} {Universalner: Targeted distillation from large language models for open named entity recognition}.

\end{thebibliography}

\newpage
\appendix

\section{Appendix}
\label{sec:appendix}
\subsection{Input Prompt}
\label{Appendix:GPT4InputPrompt}
\begin{figure*}[!t]
\centering
\includegraphics[width=14cm, height=10cm]{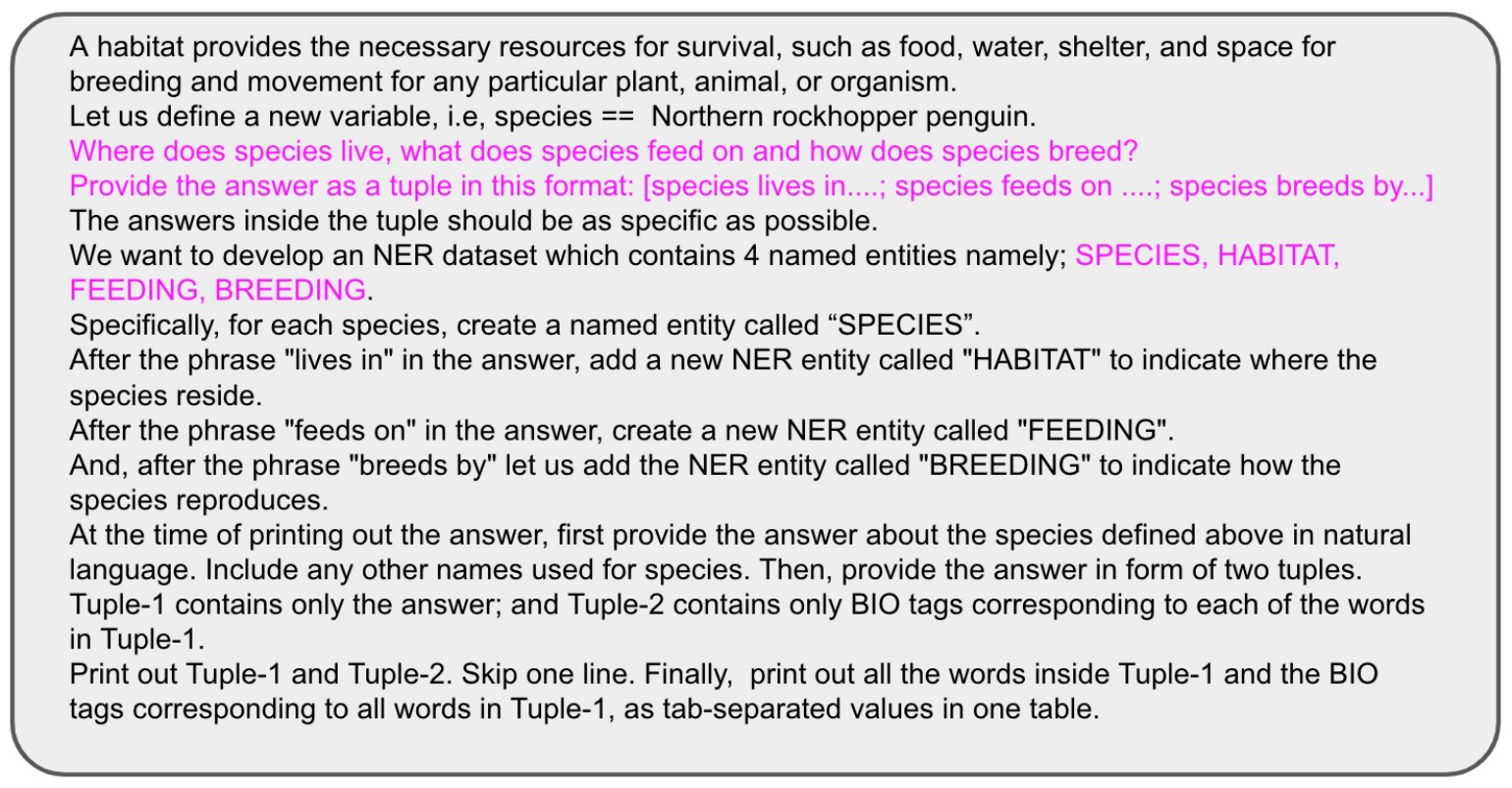}
\caption{Prompt used to generate all NER and RE data.}
\label{fig:GPT4InputPrompt}
\end{figure*}

The prompt used to generate all NER and RE data in this study is shown in Figure~\ref{fig:GPT4InputPrompt}.

\subsection{Common Names and Scientific Names}
\label{Appendix:Scientific Names}
Note that one specie may have more than one name, so we summarized the name-count in Table~\ref{Table:SpecieNameCount}. In our dataset, 85\% of species are represented by at least two names: one common name and one scientific name.
\begin{table}[!h]
\footnotesize
\centering
\begin{tabular}{lllllll}
\toprule
\textbf{\#Names} & 1  & 2  & 3 & 4 & 5 & 6 \\
\midrule
\textbf{\#Species}   & 85 & 399 & 86 & 31 & 5 & 1  \\
\bottomrule
\end{tabular}
\caption{\label{Table:SpecieNameCount}Number of names for each specie. We can see that most species in our dataset have 2 names, that is, one \texttt{common name} and one \texttt{scientific name}. }
\end{table}
\subsection{Quality of GPT-4 Output}
\label{Appendix:GPT4Quality}
We have shown details about the quality of species' information generated by GPT-4 in two tables, Table~\ref{Table:GPT4Quality} and Table~\ref{Table:AllF1Scores}.
\begin{table}[!h]
\footnotesize
\centering
\begin{tabular}{llrr}
\toprule
\textbf{Species'} & \textbf{\#Unique} & \textbf{GPT-4 Data}       & \textbf{\%Available}  \\
\textbf{Category} & \textbf{Species} & \textbf{Available}  &    \\
\midrule
Amphibians        & 153               & 86                 & 56.21 \\
Arthropods        & 150               & 74                 & 49.33 \\
Birds             & 151               & 147                & 97.35 \\
Fishes            & 153               & 109                & 71.24 \\
\midrule
\textbf{Total} &  607                 & 416                 & 68.54 \\
\bottomrule
\end{tabular}
\caption{\label{Table:GPT4Quality}We show the number of times GPT-4 had an answer for each category of species. Whenever, it did not have an answer, we explicitly ask GPT-4 to mention that ``no species information is available''.}
\end{table}
\begin{table*}[!t]
\centering
\begin{tabular}{llllllllll}
\toprule
\textbf{Category} &  \multicolumn{3}{c}{ {\bf Breeding}} &  \multicolumn{3}{c}{ {\bf Feeding} } &  \multicolumn{3}{c}{ {\bf Habitat} }\\
\textbf{of}     &&&&\\
\textbf{Species} & \textbf{P} & \textbf{R} & \textbf{F} & \textbf{P} & \textbf{R} & \textbf{F} & \textbf{P} & \textbf{R} & \textbf{F}\\
\hline
Amphibians       & 81.39 & 51.09 & 62.96     & 88.37 & 53.15 & 66.39     & 82.56 & 51.45 & 63.73\\
Arthropods       & 93.24 & 47.59 & 63.33     & 93.24 & 47.59 & 63.06     & 81.08 & 44.12 & 57.14\\
Birds            & 96.53 & 95.95 & 96.24     & 98.64 & 93.55 & 96.02     & 95.92 & 97.24 & 96.57\\
Fish             & 95.41 & 60.47 & 74.05     & 96.33 & 62.50 & 75.93     & 85.32 & 67.88 & 75.59\\
\hline
\textbf{Average} & 91.64 & 63.78 & 74.14      & 94.15 & 64.20 & 75.35     & 86.22 & 65.17 & 73.26\\
\bottomrule
\end{tabular}
\caption{\label{Table:AllF1Scores}
We measured the quality of the text generated by GPT-4, for 3 NE, by comparing it with the gold answers in external knowledge bases. We excluded the \texttt{Species} NE in this evaluation because it was part of the input prompt. All values for precision (P), recall (R) and F1-score (F) are shown in percentage (\%). GPT-4 text generated for \texttt{Birds} was of highest quality.} 
\end{table*}

\subsection{Fine-tuning BERT models}
\label{Appendix:FinetuningBERTmodels}
\begin{figure}[!t]
\centering
\includegraphics[width=6.5cm, height=3.5cm]{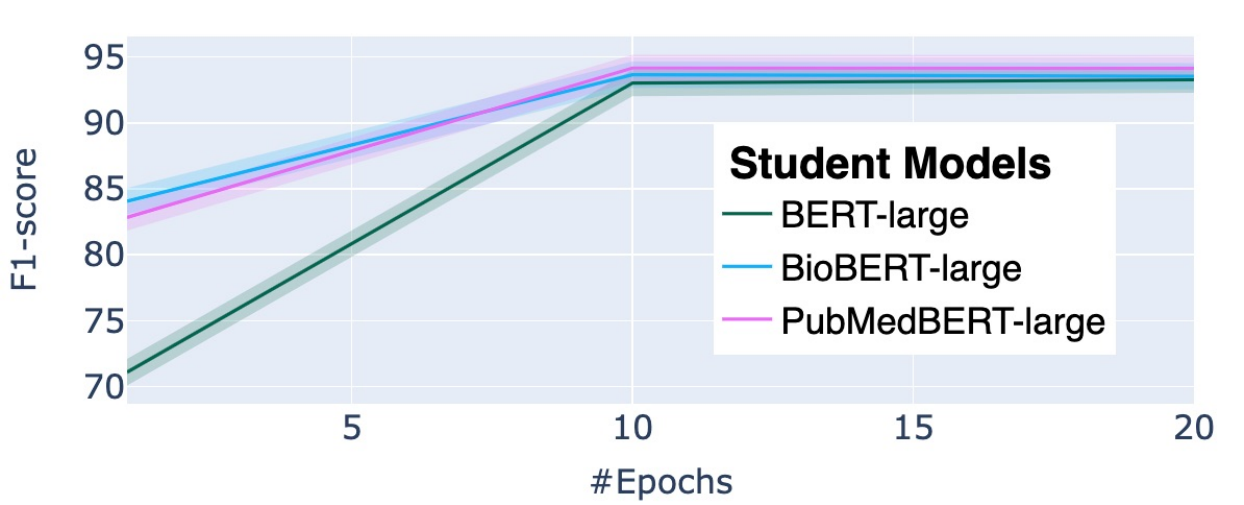}
\caption{NER performance for each student model measured by F1-scores.}
\label{fig:F1_NER_Student}
\end{figure}

Figure~\ref{fig:F1_NER_Student} indicates how BERT-large, BioBERT-large, and PubMedBERT-large performed when fine-tuned for NER in endangered species after 1, 10 and 20 epochs. 

When fine-tuned for only one epoch, there is a large gap in NER performance between general BERT and the two domain-specific BioBERT, PubMedBERT models. However, after training for 10 epochs, general BERT performance becomes comparable to both BioBERT and PubMedBERT. 

\subsection{Easy and Hard Examples}
\label{Appendix:EasyAndHardExamples}
\begin{figure}[!t]
\centering
\includegraphics[width=5.5cm, height=2.5cm]{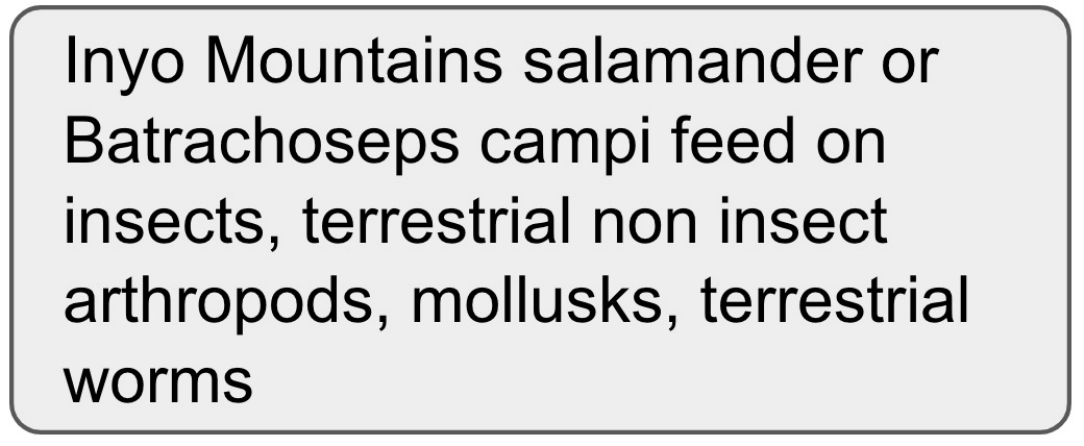}
\caption{An example of an ``easy'' text during human evaluation, easy text contains only one sentence.}
\label{fig:EasyText}
\end{figure}

\begin{figure*}[!t]
\centering
\includegraphics[width=14cm, height=14cm]{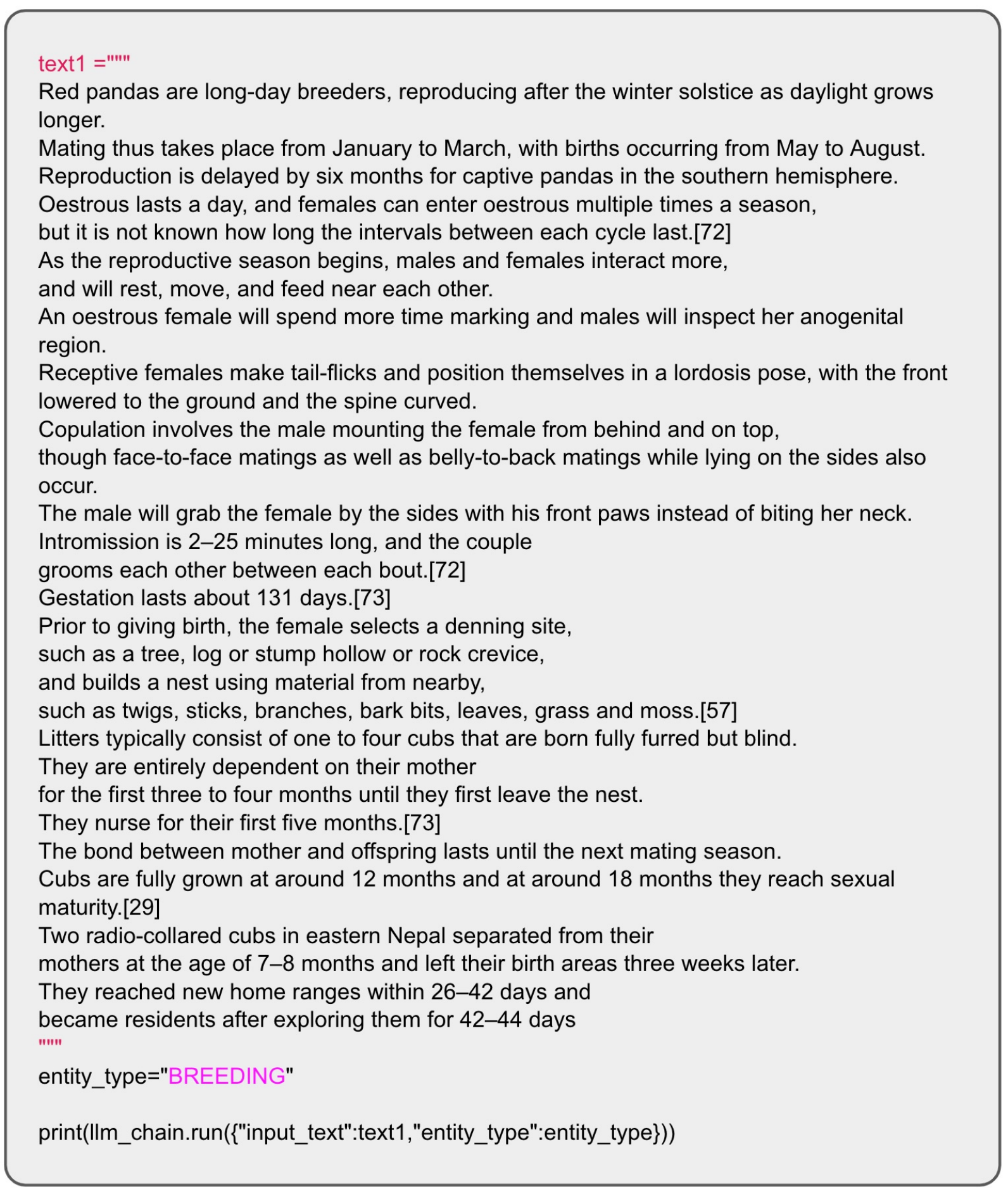}
\caption{An example of a ``hard'' text during human evaluation. Instead of adding one sentence to UniversalNER as input, we fed several paragraphs to the UniversalNER. Then we evaluated UniversalNER zero-shot ability considering \texttt{partial matches} between the gold answer and the answer provided by UniversalNER.}
\label{fig:HardText}
\end{figure*}

During zero-shot NER evaluation, we analysed the ability of ``powerful'' LLM to extract named entity information accurately from text.
We categorized the text into ``easy'' and ``hard''. Examples of both texts are shown in Figure~\ref{fig:EasyText}, and Figure~\ref{fig:HardText}. 

\end{document}